\documentclass[letterpaper, 10 pt, conference]{ieeeconf}
\IEEEoverridecommandlockouts
\usepackage{cite}
\usepackage{amsmath,amssymb,amsfonts}
\usepackage{algorithmic}
\usepackage{graphicx}
\usepackage{textcomp}
\usepackage{xcolor}
\usepackage{multirow}
\usepackage{array}
\usepackage{makecell}
\usepackage{booktabs}
\usepackage{threeparttable}

\PassOptionsToPackage{hyphens}{url}
\usepackage{xurl}
\usepackage[breaklinks=true,hidelinks]{hyperref}

\def\BibTeX{{\rm B\kern-.05em{\sc i\kern-.025em b}\kern-.08em
    T\kern-.1667em\lower.7ex\hbox{E}\kern-.125emX}}

\title{\LARGE \bf NORM-Nav: Zero-Shot Mobile Robot Navigation with Natural Language Behavioral Constraints}

\author{
Dongjie Huo\textsuperscript{1*},
Junhui Wang\textsuperscript{2,3*$\ddagger$},
Chao Gao\textsuperscript{2$\dagger$},
Yan Qiao\textsuperscript{3},
Dong Zhang\textsuperscript{1},
Guyue Zhou\textsuperscript{2,4$\dagger$}
\thanks{* Equal contribution.}
\thanks{$\ddagger$ Project lead.}
\thanks{$\dagger$ Corresponding authors.}
\thanks{
\textsuperscript{1}College of Information Science and Technology, Beijing University of Chemical Technology,
\textsuperscript{2}Institute for AI Industry Research (AIR), Tsinghua University, \textsuperscript{3}Institute of Systems Engineering and Collaborative Laboratory for Intelligent Science and Systems, Macau University of Science and Technology, 
\textsuperscript{4}School of Vehicle and Mobility, Tsinghua University.}
\thanks{Sponsored by Xinchen Qihang Inc. and XIAOMI Fund, and supported by the National Natural Science Foundation of China (No.~52475005), and in part by the Fundamental Research Funds for the Central Universities under Grant PY2505.}
}

\begin{document}
\maketitle

\begin{abstract}

Mobile robots operating in human-centered environments must generate not only collision-free paths but also trajectories that follow local behavioral conventions. Conventional costmap-based navigation emphasizes geometric feasibility and often overlooks such requirements, which can result in socially inappropriate behaviors. This paper presents NORM-Nav, a zero-shot framework that integrates natural language behavioral constraints into costmap-based planning. An LLM parses each instruction into structured constraints and grounds them using real-time vision--LiDAR perception. These constraints are encoded as multi-layer costmaps that represent geometric, semantic, directional, and velocity cues and are directly compatible with standard grid-based planners. Simulation and real-world experiments indicate that NORM-Nav improves task success rates and produces trajectories closer to human references than representative baselines. The project website is available at \url{https://ei-nav.github.io/NORM-Nav}.
\end{abstract}


\section{Introduction}

\begin{figure}[!t]
    \centering
    \includegraphics[width=0.48\textwidth]{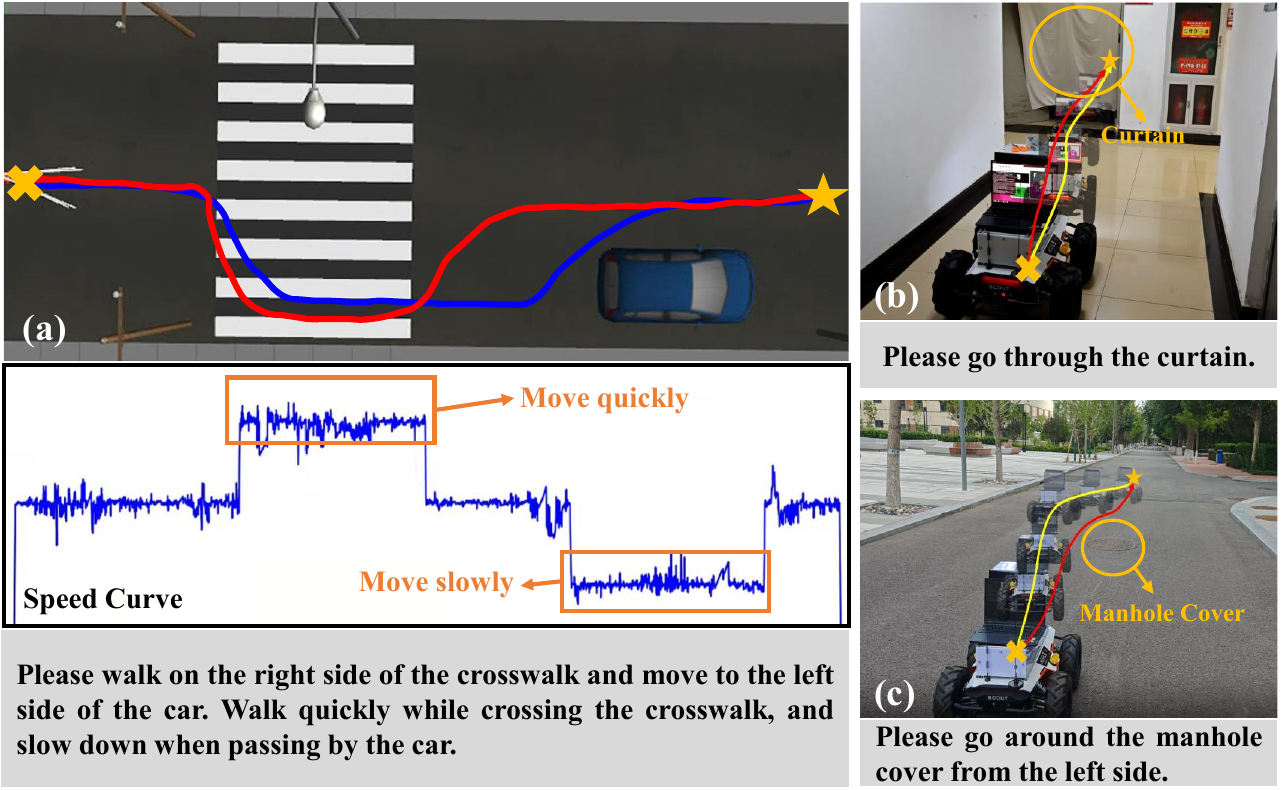}%
    \caption{Examples of robot navigation under natural language behavioral constraints. The blue line indicates the executed trajectory, while the red line shows a human-preferred reference path. The robot is able to (a) adjust its speed while complying with behavioral constraints, (b) traverse perceptually detected but traversable obstacles such as curtains, and (c) comply with side-specific instructions, such as bypassing a manhole cover from the left even though it is physically traversable.}
    \label{teaser}
    \vspace{-4mm}
\end{figure}

Mobile robots are increasingly deployed in complex and dynamic environments shared with humans. In such settings, effective navigation requires more than collision-free path planning. Robots must also conform to behavioral rules that are implicitly expected by people \cite{yue2024safe,zhang2025mpc}, such as keeping to one side of a corridor or adjusting speed when approaching vulnerable road users. While trajectories that disregard these conventions may remain geometrically feasible, they often appear unnatural, socially inappropriate, or unsafe, thereby limiting the acceptance of robots in human environments. Representative examples of such behavioral expectations are illustrated in Fig.~\ref{teaser}.

Costmap-based navigation frameworks continue to serve as the backbone of most deployed systems due to their computational efficiency and robustness \cite{wang2025openbench, Wang2025OPEN}. These approaches typically represent the environment as a grid in which all free space is treated as traversable. As a result, they fail to distinguish between semantically different regions, for example, sidewalks and grassy areas, and cannot adapt to context-specific behavioral requirements \cite{macenski2020marathon}. Furthermore, reliance on purely geometric perception can result in misinterpretation. Tall grass may be treated as an obstacle, whereas lightweight curtains at a doorway may be mistakenly perceived as rigid barriers. Such limitations reduce efficiency and prevent robots from exhibiting socially compliant behaviors.

Learning-based approaches have been investigated to alleviate these challenges \cite{Lin2025AdvancesEmbodiedNavigation,Wu2023VisionLanguageSurvey,Zhang2024VisionLanguageSurvey}. These methods directly map natural language and sensory input to low-level actions, providing flexibility in principle. However, they often depend on large-scale datasets that are costly to collect and seldom capture the diversity of real-world environments. Consequently, they tend to generalize poorly to unseen scenarios and may exhibit brittle or opaque behaviors. Reinforcement and imitation learning strategies offer richer modeling of behaviors \cite{Weerakoon2022TERP,Jiang2024BEVNav}, but require substantial training effort and remain sensitive to distribution shifts, which constrains their practicality.

Large language models (LLMs) can parse natural language instructions and support rule-related reasoning without task-specific training \cite{Minaee2025LLMSurvey,Shah2023LMNav}. Combined with camera--LiDAR perception, they can translate free-form instructions into structured navigation constraints \cite{weerakoon2025behav}. This enables integrating user intent into planning while retaining the modularity of classical navigation frameworks.

This work proposes \textit{NORM-Nav}, a zero-shot navigation framework that integrates natural language behavioral constraints into costmap-based planning. NORM-Nav converts each instruction into structured constraints (spatial preference, velocity adjustment, and traversability) and grounds them through real-time vision--LiDAR fusion. The constraints are encoded as multi-layer costmaps that are directly consumable by standard grid-based planners.

We evaluate NORM-Nav in simulation and real-world experiments. Relative to representative baselines, NORM-Nav achieves higher success rates, follows user-specified behavioral constraints more consistently, and produces trajectories that are closer to human references.

The main contributions are summarized as follows:
\begin{itemize}
    \item We propose NORM-Nav, a zero-shot navigation framework that translates natural language behavioral constraints into structured planning cues and integrates them into costmap-based navigation.
    \item We present a modular, plug-in design that augments standard costmap-based stacks without changes to the underlying planner.
    \item We demonstrate the effectiveness of NORM-Nav through  simulation and real-world experiments.
\end{itemize}

\section{Related Work}

\subsection{Semantic Extensions to Classical Navigation}
Costmap-based frameworks remain the cornerstone of classical robot navigation, where the environment is discretized into free and occupied cells \cite{Zheng2021ROSTuningBook}. While computationally efficient and widely deployed, such representations cannot account for semantic or socially motivated preferences. To address this limitation, researchers have proposed augmenting costmaps with semantic information \cite{sani2024improving, mao2025pacer}. By embedding contextual cues, robots are able to differentiate between regions that are geometrically alike but behaviorally distinct. For example, Zhang et al.~\cite{zhang2024interactive} combined an LLM with a vision–language model (VLM) to reinterpret LiDAR returns, allowing objects such as curtains to be considered traversable. While these techniques demonstrate the benefit of linking semantics with navigation, they are difficult to generalize to broader behavioral norms. 
Instead of embedding fixed semantic priors into costmaps, we enable flexible interpretation of semantic instructions through a structured integration with classical navigation, thereby retaining robustness while extending behavioral variety.

\subsection{Language-Driven Navigation}
End-to-end vision–language navigation (VLN) methods map user instructions and sensory observations into motion policies through a unified model \cite{liu2024volumetric, zhang2024navid}. These approaches achieve competitive performance in benchmark evaluations \cite{Lin2025AdvancesEmbodiedNavigation, Wu2023VisionLanguageSurvey, Zhang2024VisionLanguageSurvey}, but real-world deployment remains challenging. The main difficulties include dependence on large-scale annotated datasets \cite{wang2023scaling}, limited generalization to unseen environments \cite{hirose2024lelan}, and fragile behavior under perceptual uncertainties \cite{glossop2025cast}.

Zero-shot methods attempt to address these limitations by leveraging pretrained LLMs and VLMs to interpret instructions without task-specific supervision \cite{yokoyama2024vlfm, dorbala2022clip, ahn2022can}. Natural language is converted into structured representations and aligned with open-vocabulary visual recognition \cite{zhou2024navgpt2, kuang2024openfmnav, long2024instructnav}, thereby supporting flexible generalization. Nevertheless, their low decision frequency often hinders reliable deployment in realistic settings. Hybrid methods have therefore been proposed to combine semantic reasoning with traditional planning. BehAV \cite{weerakoon2025behav} integrates language-specified constraints into classical planning, enhancing interpretability. Our method follows a similar philosophy but adopts a more generalized formulation. Instead of tightly coupling language constraints to planning rules, we introduce a lightweight semantic reasoning module that maps free-form instructions onto structured navigation primitives. This design preserves the efficiency and reliability of costmap-based navigation, while offering broader adaptability to diverse behavioral instructions.

\begin{figure*}[t!] 
        \centering
        \includegraphics[width=\textwidth]{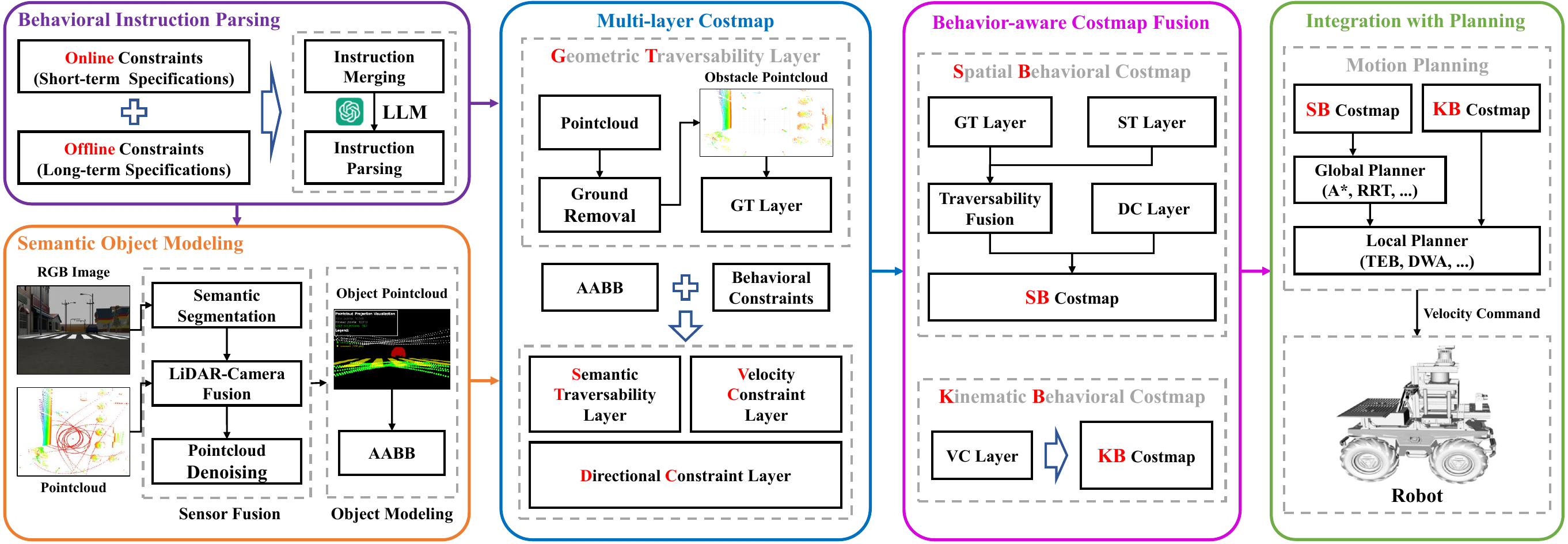}
        \caption{The architecture of the proposed method for zero-shot navigation under natural language behavioral constraints. The system integrates LLMs with vision–LiDAR perception and encodes parsed behavioral instructions into multi-layer costmaps for motion planning.}
        \label{system_framework}        
        \vspace{-4mm}
\end{figure*}

\section{Problem Formulation}
This work considers the problem of robot path planning under natural language behavioral constraints with real-time perception. The robot is placed in an environment with a start position $S$ and a goal position $T$, and is provided with a set of behavioral constraints $\mathcal{C}$ stated in natural language. Examples include instructions such as ``avoid walking on the grass,'' ``keep to the right side of the road,'' or ``pass vehicles on the left.'' The objective is to generate a feasible trajectory $\tau$ from $S$ to $T$ that satisfies all constraints in $\mathcal{C}$.

During navigation, the robot continuously acquires multimodal sensory observations $O_t$ from its onboard LiDAR and camera at time step $t$. These data are essential for detecting obstacles, identifying semantic elements, and associating the behavioral constraints with the current environment. As the robot progresses, the planned trajectory must be adaptively updated based on the latest perception to ensure that all specified constraints are respected.

Formally, let $\mathcal{P}$ denote the set of all feasible trajectories from $S$ to $T$ that are consistent with the robot’s motion capabilities and the observed environment $\{O_t\}$. The problem is to find an optimal trajectory
\begin{equation}
\tau^* = \arg\min_{\tau \in \mathcal{P}} J(\tau)
\quad \text{subject to} \quad \tau \models \mathcal{C},
\end{equation}
where $J(\tau)$ is a navigation cost function, and $\tau \models \mathcal{C}$ indicates that $\tau$ fulfills all constraints in $\mathcal{C}$ after they are interpreted and grounded through real-time sensory perception.

In summary, the inputs to the problem consist of the navigation pair $(S, T)$, the sensory stream $\{O_t\}$, and the natural language constraints $\mathcal{C}$, while the output is an executable trajectory $\tau^*$ that minimizes $J(\tau)$ and satisfies $\mathcal{C}$. The key challenge lies in transforming inherently ambiguous natural language instructions into reliable, perception-grounded constraints and in ensuring that the derived trajectory remains both feasible and compliant as the environment evolves.

\section{Methodology}
This work proposes a zero-shot navigation framework that enables robots to interpret and execute behavioral constraints expressed in natural language without relying on additional task-specific training. The core idea is to integrate LLMs with conventional costmap-based navigation so that planned trajectories remain both geometrically feasible and behaviorally appropriate. Instead of relying solely on geometric representations, the framework parses long-term rules and short-term situational instructions into structured constraints, grounds them through real-time fusion of visual and LiDAR perception, and encodes them within the cost space. These constraints are organized into multi-layer costmaps that incorporate geometric, semantic, spatial, and kinematic information, which are subsequently fused to guide planning. The overall system architecture is illustrated in Fig.~\ref{system_framework}.

\subsection{Behavioral Instruction Parsing}
\begin{figure}[!t]
    \centering
    \includegraphics[width=0.48\textwidth]{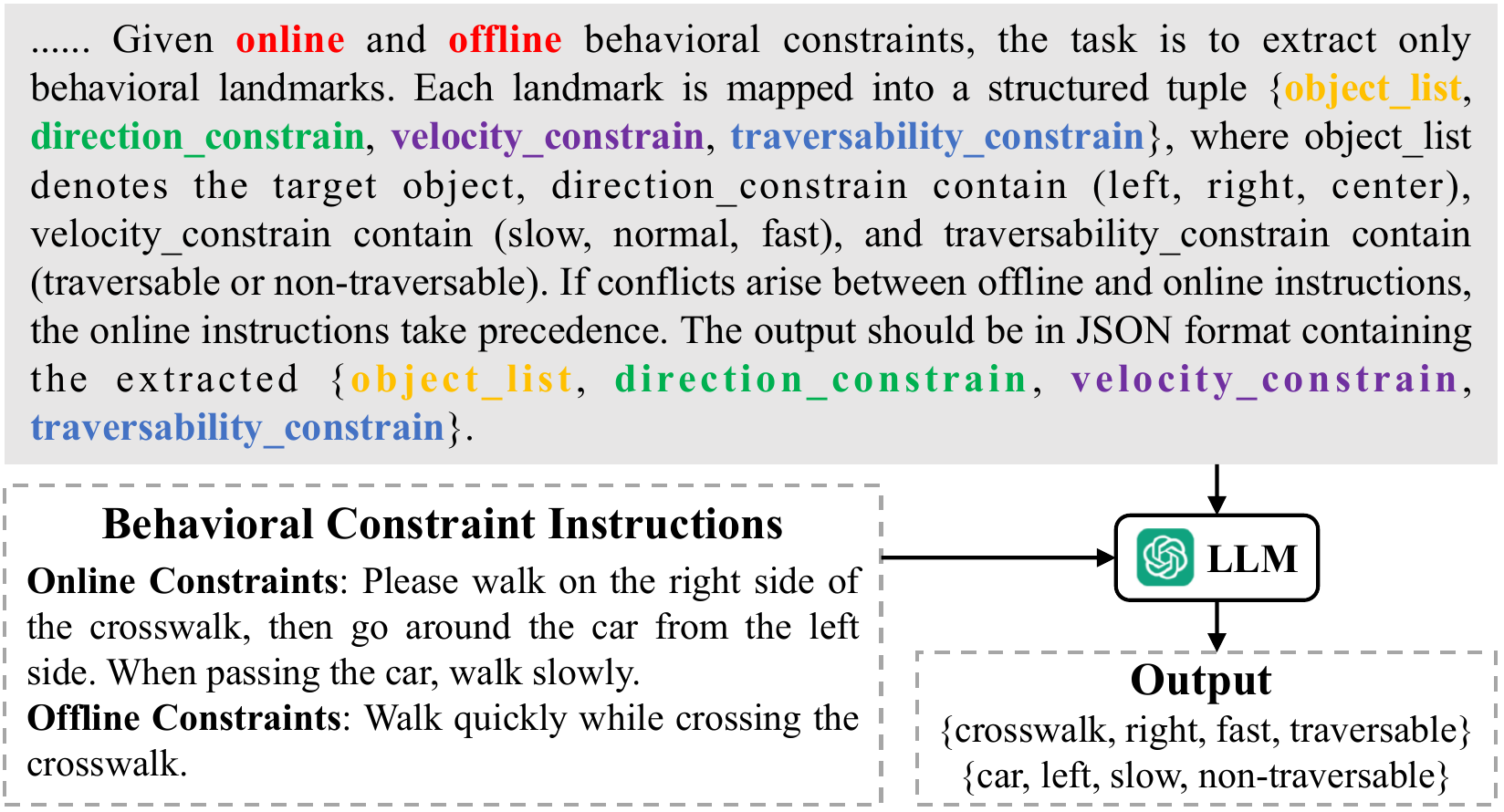}%
    \caption{Example of parsing online and offline behavioral constraints into structured representations.}
    \label{behavior_parsing}
    \vspace{-4mm}
\end{figure}
Natural language provides a direct and flexible interface for humans to specify navigation behaviors.  
To accommodate instructions with different temporal scopes, behavioral constraints are categorized into two groups.

\subsubsection{Offline constraints}  
The set of offline behavioral constraints is defined as
\begin{equation}
\mathcal{L}_{off} = \{ L^{off}_1, L^{off}_2, \dots, L^{off}_m \},
\end{equation}
where each $L^{off}_i$ encodes global requirements that remain valid for the entire task, such as ``\textit{keep right when walking on the road}'' or ``\textit{avoid entering the grass}.''

\subsubsection{Online constraints}  
The set of online behavioral constraints is defined as
\begin{equation}
\mathcal{L}_{on} = \{ L^{on}_1, L^{on}_2, \dots, L^{on}_n \},
\end{equation}
where each $L^{on}_j$ specifies short-term requirements triggered by particular situations, such as ``\textit{slow down near the car}'' or ``\textit{pass through the curtain}.''

\subsubsection{Unified constraints}  
For subsequent processing, the two sets are merged into a unified specification:
\begin{equation}
\mathcal{L} = \mathcal{L}_{off} \cup \mathcal{L}_{on}.
\end{equation}

This specification is then parsed by an LLM into a structured representation, as illustrated in Fig. \ref{behavior_parsing}.
\begin{equation}
\mathcal{C} = \{(O_i, d_i, v_i, t_i) \mid i = 1, \ldots, r \},
\end{equation}
where $O_i$ denotes the referenced object, $d_i$ encodes spatial preference, $v_i$ indicates velocity requirements, and $t_i$ represents semantic traversability.  
Through this process, both long-term norms and short-term situational instructions are cast into a unified symbolic form that can guide perception and planning.

Table~\ref{tab:constraint_def} summarizes the definition and representative values of each element.  
The LLM interprets free-form natural language instructions and maps them into these representative values, thereby producing consistent structured tuples.

\begin{table*}[t]
  \centering
  \caption{Definition and Representative Values for Each Element in Structured Behavioral Constraints}
  \label{tab:constraint_def}
  \begin{tabular}{c c c c}
    \toprule
    \textbf{Symbol} & \textbf{Meaning} & \textbf{Representative Values} & \textbf{Example Instruction} \\
    \midrule
    $O_i$ & Referenced object & pedestrian, vehicle, grass, crosswalk, \ldots & ``Walk on the crosswalk'', ``Avoid the grass'' \\
    $d_i$ & Directional preference & \{$\emptyset$, left, right, middle\} & ``Pass the car on the left'', ``Keep to the right side of the road'' \\
    $v_i$ & Velocity requirement & \{$\emptyset$, slow, normal, fast\} & ``Walk slowly near the car'', ``Move quickly when crossing the street'' \\
    $t_i$ & Semantic traversability & \{$\emptyset$, traversable, non-traversable\} & ``Go through the curtain'', ``Lawn is non-traversable'' \\
    \bottomrule
  \end{tabular}
  \vspace{-2mm}
\end{table*}

\begin{table}[t]
\centering
\caption{Assignment of $(c_1,c_2,c_3)$ under different directional constraints}
\label{tab:spatial-rules}
\begin{tabular}{c c c c}
\toprule
Constraint & $c_1$ & $c_2$ & $c_3$ \\
\midrule
left   & $c_{\min}$ & $\tfrac{1}{2}(c_{\min}+c_{\max})$ & $c_{\max}$ \\
right  & $c_{\max}$ & $\tfrac{1}{2}(c_{\min}+c_{\max})$ & $c_{\min}$ \\
middle & $c_{\max}$ & $c_{\min}$ & $c_{\max}$ \\
\bottomrule
\end{tabular}
\vspace{-5mm}
\end{table}

\subsection{Semantic Object Modeling}
To interpret behavioral constraints within the observed environment, semantic object models are constructed via camera–LiDAR fusion in two stages: open-vocabulary semantic grounding and multi-frame aggregation.

\subsubsection{Open-vocabulary semantic grounding}
Given an RGB image $I_t \in \mathbb{R}^{H \times W \times 3}$ at time step $t$ and the textual description of an object $O_i$, GSAM2~\cite{ren2024grounded} generates a binary mask $M_i$. The pixel set corresponding to $O_i$ is
\begin{equation}
S_i = \{ (u, v) \mid M_i(u,v) = 1 \}
\end{equation}
The LiDAR point cloud at time $t$ is denoted as $P_t = \{p_j=(x_j, y_j, z_j)\}_{j=1}^N$. Each point $p_j$ is projected to the image plane via the projection matrix and is associated with $O_i$ if its projection lies in $S_i$, yielding the object-specific subset $P_i^t \subseteq P_t$.

\subsubsection{Multi-frame aggregation}
To enhance robustness, object models are aggregated over a sliding temporal window. At each time step $t$, $P_i^t$ is refined using DBSCAN \cite{deng2020dbscan} to suppress noise, resulting in $\hat{P}_i^t$. The temporally aggregated model is
\begin{equation}
P_i(t) = \bigcup_{\tau = t-T+1}^{t} \hat{P}_i^\tau
\end{equation}
where $T$ is the aggregation horizon. On the bird’s-eye view (BEV) plane, object $O_i$ is approximated by an axis-aligned bounding box (AABB) given by
\begin{equation}
B_i = \left( \min_{p \in P_i}(x,y), \; \max_{p \in P_i}(x,y) \right)
\end{equation}
This BEV-projected bounding box provides a compact representation for encoding behavioral constraints in costmap construction.

\subsection{Multi-layer Costmap with Semantic Guidance}
Trajectories consistent with natural language constraints are generated by constructing a multi-layer costmap. Four complementary layers are considered: geometric traversability, semantic traversability, directional constraints, and velocity constraints.

\subsubsection{Geometric Traversability Layer}
This layer encodes free space and obstacle occupancy. From $P_t$, ground points and points above the maximum robot height $h_{\max}$ are removed. The remaining obstacles form
\begin{equation}
P^{obs}_t = \{ p_j \mid z_{\min} < z_j < h_{\max}, \; p_j \notin P^{ground}_t \}
\end{equation}
where $P^{ground}_t$ denotes estimated ground. Each obstacle is projected onto a BEV grid to yield
\begin{equation}
C_{geo}(u,v) =
\begin{cases}
255, & \exists p_j \in P^{obs}_t \;\; \text{s.t. }\pi_{\text{BEV}}(p_j)=(u,v) \\
100, & \text{otherwise}
\end{cases}
\end{equation}
with $\pi_{\text{BEV}}(\cdot)$ the projection function.

\subsubsection{Semantic Traversability Layer}
Semantic information from $\mathcal{C}$ is incorporated. For object $B_i$ with semantic traversability $t_i$, the average geometric cost is
\begin{equation}
\bar{C}_{geo}(B_i) = \frac{1}{|B_i|} \sum_{(u,v)\in B_i} C_{geo}(u,v)
\end{equation}
If $t_i = \emptyset$, it is inferred as
\begin{equation}
t_i =
\begin{cases}
\text{non-traversable}, & \bar{C}_{geo}(B_i) > \tau \\
\text{traversable}, & \text{otherwise}
\end{cases}
\end{equation}
The semantic traversability cost is then defined as
\begin{equation}
C_{sem}(u,v) =
\begin{cases}
1, & (u,v)\in B_i, \; t_i=\text{traversable} \\
2, & (u,v)\in B_i, \; t_i=\text{non-traversable} \\
0, & \text{otherwise}
\end{cases}
\end{equation}
This step allows dynamic regulation of traversable regions based on semantic priors.

\subsubsection{Directional Constraint Layer}
Directional constraints are employed to encode side-specific navigation preferences with respect to semantic objects. When an instruction specifies $d_i \neq \emptyset$, the bounding box $B_i$ of the corresponding object is first enlarged by a margin $d$ if $t_i = \text{non-traversable}$, so that the constraint also applies around the obstacle boundary for safe planning.

\begin{figure}[!t]
    \centering
    \includegraphics[width=0.3\textwidth]{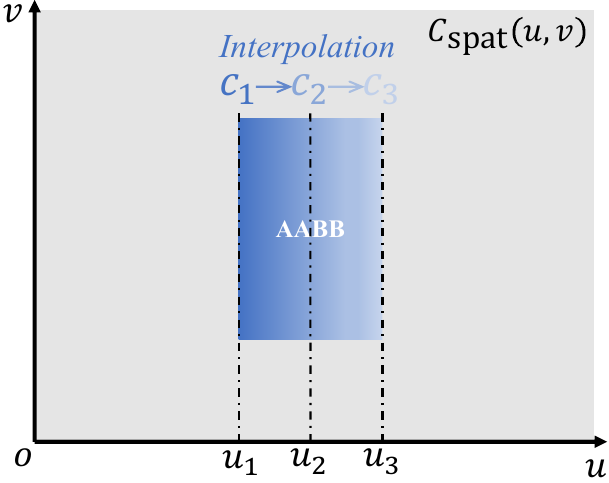}%
    \caption{Illustration of the directional constraint layer.}
    \label{interp}
    \vspace{-4mm}
\end{figure}

For cost assignment, three boundary values $(c_1, c_2, c_3)$ are placed along the lateral axis of $B_i$, with their values determined by the directional rule (see Table~\ref{tab:spatial-rules}). As illustrated in Fig.~\ref{interp}, the bounding box is divided into two consecutive regions. The cost distribution is obtained through piecewise interpolation:  
\begin{itemize}
    \item from $c_1$ to $c_2$ across the first region,  
    \item from $c_2$ to $c_3$ across the second region.  
\end{itemize}
Both regions are interpolated using the same function. Let $u$ denote the horizontal coordinate of a grid cell. The interpolation function is expressed as
\begin{equation}
\label{eq:interpolation}
C(u) =
\begin{cases}
\left\lfloor c_{1} + (c_{2} - c_{1}) \left(\dfrac{u-u_{1}}{u_{2}-u_{1}}\right)^{\alpha} \right\rfloor, & u \in [u_{1}, u_{2}], \\[8pt]
\left\lfloor c_{2} + (c_{3} - c_{2}) \left(\dfrac{u-u_{2}}{u_{3}-u_{2}}\right)^{\alpha} \right\rfloor, & u \in [u_{2}, u_{3}],
\end{cases}
\end{equation}
where $(u_{1},u_{2},u_{3})$ denote the left boundary, midpoint, and right boundary of $B_i$, and the parameter $\alpha>0$ controls the sharpness of the gradient. When $\alpha=1$, the interpolation is linear. 
The operator $\lfloor \cdot \rfloor$ ensures that the interpolated costs are discretized to integer values for consistency with the grid-based costmap.
The preferred side specified by the instruction corresponds to the low-cost region, whereas the non-preferred side is penalized by higher costs. The resulting cost field allows standard planners to naturally generate trajectories that adhere to side-specific navigation preferences.

\subsubsection{Velocity Constraint Layer}\label{vel_layer}
Velocity rules are encoded when $v_i \neq \emptyset$. The resulting cost field is
\begin{equation}
C_{vel}(u,v) =
\begin{cases}
c_{\min}, & (u,v)\in B_i,\; v_i=\text{slow} \\[6pt]
\dfrac{1}{2}\,(c_{\min}+c_{\max}), & (u,v)\in B_i,\; v_i=\text{normal} \\[6pt]
c_{\max}, & (u,v)\in B_i,\; v_i=\text{fast} \\[6pt]
0, & \text{otherwise}
\end{cases}
\end{equation}

\subsection{Behavior-aware Costmap Fusion}
The four layers are combined into two categories of behavioral costmaps: a spatial behavioral costmap $C_{spat}$, derived from geometric, semantic, and directional constraints; and a kinematic behavioral costmap $C_{vel}$, derived from velocity preferences.

\subsubsection{Spatial Behavioral Costmap}
Geometric and semantic traversability layers are first fused:
\begin{equation}
\begin{aligned}
I(u,v) =\;& \tfrac{1}{2} \, C_{geo}\odot(\mathbf{1}-C_{sem})(\mathbf{2}-C_{sem}) \\
&+ 100 \cdot C_{sem}\odot(\mathbf{2}-C_{sem}) \\
&+ \tfrac{1}{2}\cdot 255 \cdot C_{sem}\odot(C_{sem}-\mathbf{1})
\end{aligned}
\end{equation}
where $\odot$ denotes element-wise multiplication and $\mathbf{1}, \mathbf{2}$ are matrices of all ones and all twos. This step corrects LiDAR perception results using semantic input. The result $I$ is further combined with $C_{dir}$:
\begin{equation}
\begin{aligned}
C_{spat}(u,v) &= \mathbf{1}(I=255)\odot 255 \\
&+ \mathbf{1}(I\neq 255)\odot \mathbf{1}(C_{dir}>0)\odot C_{dir} \\
&+ \mathbf{1}(I\neq 255)\odot \mathbf{1}(C_{dir}=0)\odot I
\end{aligned}
\end{equation}
where $\mathbf{1}(\cdot)$ is an indicator function.

\subsubsection{Kinematic Behavioral Costmap}
The kinematic behavioral costmap $C_{kin}$ is directly constructed from $C_{vel}$ (Sec.~\ref{vel_layer}) and encodes localized velocity modulation without further fusion.

Together, the spatial behavioral costmap $C_{\text{spat}}$ and the kinematic costmap $C_{\text{kin}}$ provide complementary guidance. The spatial term enforces geometric and semantic traversability as well as directional preferences, whereas the kinematic term regulates local velocity. The planner then leverages these fused costmaps to generate trajectories that are collision-free and aligned with natural language behavioral specifications.

\subsection{Integration with Planning}

The fused behavioral costmaps are subsequently utilized by the motion planning to generate executable trajectories. The spatial behavioral costmap and the kinematic behavioral costmap play complementary roles. Each contributes a distinct function, and together they ensure that the robot follows paths that are both geometrically feasible and consistent with behavioral specifications provided in natural language.

\subsubsection{Spatial Behavioral Costmap for Geometric Optimization}  
The spatial behavioral costmap $C_{\text{spat}}$ can be directly applied to planners that operate on a grid-based representation of the environment. In this configuration, $C_{\text{spat}}$ augments the standard geometric costmap so that behavioral requirements, including keeping to a specific side of a road or avoiding restricted regions, are incorporated into the optimization process. Because $C_{\text{spat}}$ maintains compatibility with existing costmap-based frameworks, no modification to the underlying planning algorithms is required. This modular integration preserves generality while embedding behavior awareness into the trajectory generation process.

\subsubsection{Kinematic Behavioral Costmap for Velocity Regulation}  
The kinematic behavioral costmap $C_{\text{kin}}$ provides a complementary function by regulating the velocity of the robot during execution. Instead of influencing the spatial structure of the trajectory, $C_{\text{kin}}$ specifies speed constraints for the cell currently occupied by the robot. The cost value at the corresponding grid location is mapped into an admissible limit on translational velocity, which is forwarded to the local planner. In this way, the robot is able to adjust its speed dynamically according to user instructions, for example, slowing down near vehicles or increasing velocity when crossing a crosswalk.  

In combination, the two costmaps provide a unified mechanism for integrating behavioral constraints with conventional planning algorithms. The spatial behavioral costmap guides the generation of paths that adhere to traversability and directional preferences, while the kinematic behavioral costmap ensures that execution speed respects context-dependent motion requirements. Together, these mechanisms enable conventional planning algorithms to produce trajectories that are both socially acceptable and context aware.

\section{Experimental Results}
\subsection{Experimental Setup}
We conduct experiments in simulation and real-world environments to evaluate the effectiveness of the proposed method.

\subsubsection{Simulation Environment}
The MEDIUM environment from OpenBench~\cite{wang2025openbench} is used to examine navigation performance under various behavioral constraints. The simulated platform is a four-wheeled differential-drive robot equipped with a Livox MID-360 LiDAR and an RGB-D camera.

\subsubsection{Real-world Platform}
\begin{figure}[!t]
    \centering
    \includegraphics[width=0.40\textwidth]{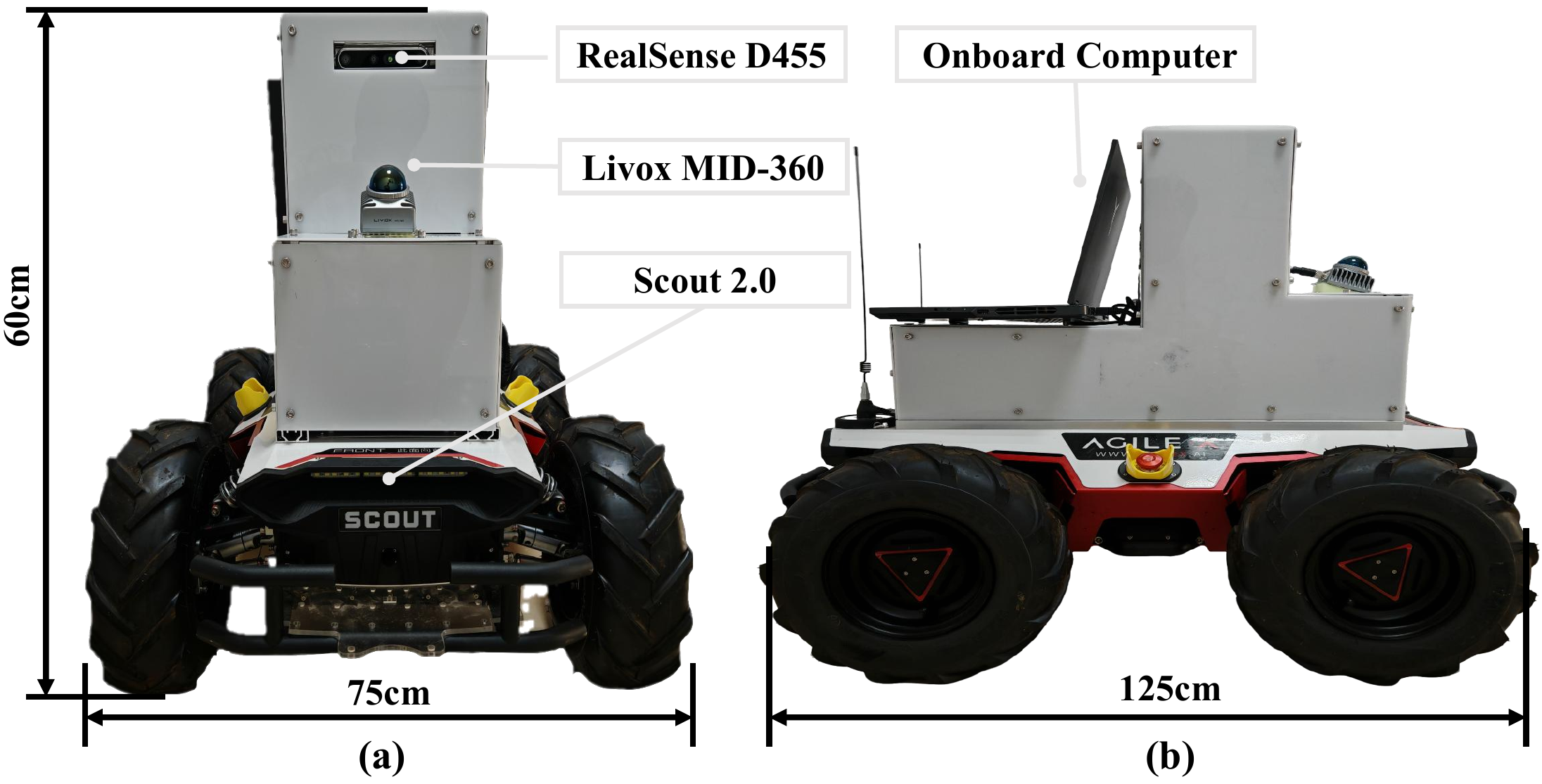}%
    \caption{The experimental robot platform: (a) front view and (b) side view.}
    \label{robot_platform}
    \vspace{-4mm}
\end{figure}
For physical validation, we employ a SCOUT~2.0 differential-drive mobile robot (Fig.~\ref{robot_platform}). The platform is equipped with a Livox MID-360 LiDAR for 3D perception, an Intel RealSense D455 depth camera for RGB-D sensing, and an onboard computer (AMD R9-7945HX CPU, NVIDIA RTX~4060 GPU) for real-time processing. The LiDAR is tilted at an angle of 20 degrees to scan the ground.

\subsubsection{Evaluation Metrics}
We assess navigation performance using four adopted metrics:
\begin{itemize}
    \item \textit{Success Rate (SR)}: the proportion of successful trials in which the robot reaches the goal while fully respecting all behavioral constraints.
    \item \textit{Success weighted by Path Length (SPL)~\cite{Anderson2018EvalNav}}: a combined measure of efficiency that reflects both success rate and path optimality relative to the shortest feasible trajectory.
    \item \textit{Fréchet Distance (FD)~\cite{aronov2006frechet}}: the similarity between the executed trajectory and a human teleoperated reference path, where smaller values indicate closer alignment to human-preferred navigation.
    \item \textit{Behavioral Following Accuracy (BFA)~\cite{weerakoon2025behav}}: the fraction of the executed trajectory length that is consistent with the specified behavioral constraints.
\end{itemize}

\subsubsection{Implementation Details}
Behavioral instructions are parsed using \textit{Qwen-VL-Max}. 
In costmap construction, both the spatial and kinematic constraint layers employ 
$c_{\min} = 0$ and $c_{\max} = 255$, with the interpolation parameter in Eq.~\ref{eq:interpolation} fixed at 
$\alpha = 3.0$. The costmap resolution is set to $0.05\,\text{m}$ 
over a horizon of $10\,\text{m} \times 10\,\text{m}$. For path generation, the global planner adopts the A* algorithm, while local trajectory optimization is performed using the Timed Elastic Band (TEB) planner.

\begin{figure*}[t!] 
        \centering
        \includegraphics[width=\textwidth]{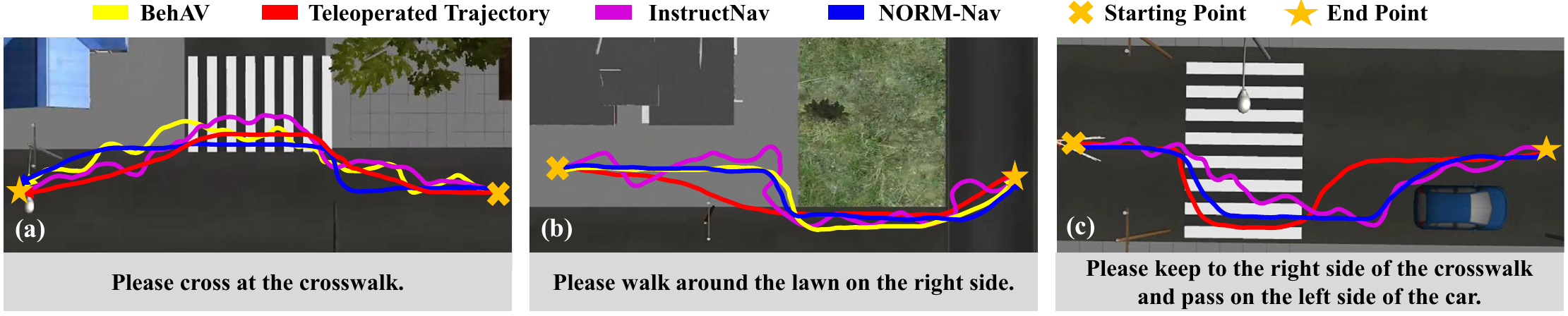}
        \caption{Simulation results on three representative navigation tasks. The proposed method produces stable trajectories that closely follow human-operated reference paths, outperforming baseline approaches.}
        \label{sim_exp}        
\end{figure*}

\subsection{Evaluation of Navigation Performance}
We evaluate the proposed method against representative language-driven navigation approaches, including BehAV~\cite{weerakoon2025behav}, InstructNav~\cite{long2024instructnav}, and the interactive navigation framework (INF)~\cite{zhang2024interactive}. All task instructions are expressed in natural language and are divided into four categories:

\begin{itemize}
    \item \textit{Region-Following Task:} Robots are required to approach or traverse designated semantic regions while following specified behavioral constraints.
    \item \textit{Region-Avoidance Task:} Robots are required to actively avoid certain semantic regions or obstacles while respecting the given behavioral constraints.
    \item \textit{Traversable-Obstacle Task:} Robots are required to navigate through regions that conventional LiDAR-based systems detect as obstacles but that are in fact traversable.
    \item \textit{Combined Task:} Robots are required to perform all of the above simultaneously, following designated regions and avoiding restricted areas while satisfying behavioral constraints.
\end{itemize}

\begin{table}[t]
\centering
\caption{Evaluation of Navigation Performance}
\label{navperformance}
\begin{tabular}{cccccc}
\toprule
\textbf{Task} & \textbf{Method} & \textbf{SR ($\uparrow$)} & \textbf{SPL ($\uparrow$)} & \textbf{FD ($\downarrow$)} & \textbf{BFA ($\uparrow$)} \\
\midrule
\multirow{4}{*}{\makecell[c]{Region \\ Following \\ Task}} 
& BehAV           & 60\% & 19.63\% & 4.21 & 62\% \\
& InstructNav     & 50\% & 15.14\% & 5.36 & 58\% \\
& INF             & --   & -- & -- & -- \\
& NORM-Nav        & \textbf{90\%} & \textbf{65.77\%} & \textbf{2.14} & \textbf{89\%} \\
\cmidrule{1-6}
\multirow{4}{*}{\makecell[c]{Region \\ Avoidance \\ Task}} 
& BehAV           & 60\% & 38.34\% & 2.82 & 64\% \\
& InstructNav     & 40\% & 17.68\% & 5.10 & 61\% \\
& INF             & -- & -- & -- & -- \\
& NORM-Nav        & \textbf{90\%} & \textbf{58.92\%} & \textbf{2.16} & \textbf{87\%} \\
\cmidrule{1-6}
\multirow{4}{*}{\makecell[c]{Traversable \\ Obstacle \\ Task}} 
& BehAV           & 0\% & -- & -- & -- \\
& InstructNav     & 0\% & -- & -- & -- \\
& INF             & 70\% & 52.14\% & 2.23 & 67\% \\
& NORM-Nav        & \textbf{80\%} & \textbf{67.26\%} & \textbf{1.35} & \textbf{84\%} \\
\cmidrule{1-6}
\multirow{4}{*}{\makecell[c]{Combined \\ Tasks}} 
& BehAV           & 30\% & 18.62\% & 7.02 & 39\% \\
& InstructNav     & 20\% & 12.21\% & 6.45 & 36\% \\
& INF             & -- & -- & -- & -- \\
& NORM-Nav        & \textbf{90\%} & \textbf{54.87\%} & \textbf{3.01} & \textbf{85\%} \\
\bottomrule
\end{tabular}
\vspace{-2mm}
\end{table}

\begin{figure*}[t!] 
    \centering
    \includegraphics[width=\textwidth]{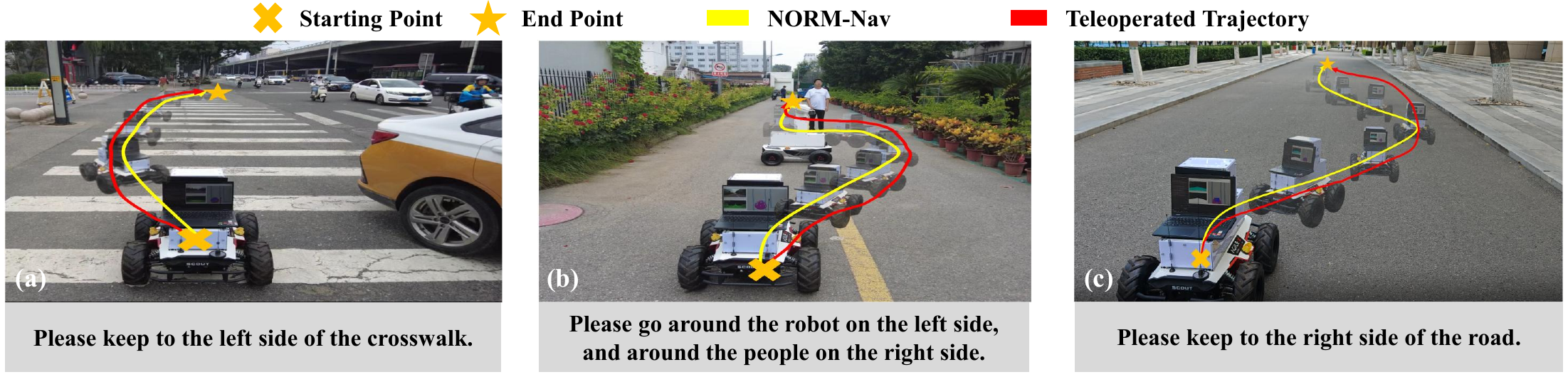}
    \caption{Real-world demonstrations of behavior-constrained navigation. The proposed method successfully follows natural language instructions without collisions, generating stable trajectories that remain close to human-operated reference paths.}
    \label{real_exp}        
    \vspace{-4mm}
\end{figure*}

In all tasks, success is defined only when the robot reaches the goal while fully complying with the given instructions. The results are summarized in Table~\ref{navperformance}, and the performance of each method is analyzed across the four task types.

In the region-following task, NORM-Nav achieves the best performance across all metrics. It follows instructions such as moving along a specified region or keeping to one side with high accuracy. The relatively small FD indicates close alignment with human-operated trajectories, while the high SPL shows improved efficiency. BehAV, relying only on VLM-based reasoning, completes some tasks but with inefficient paths. InstructNav depends on landmark-based navigation and frequently fails with large landmarks such as crosswalks. Both baselines deviate substantially from human-like trajectories when side-specific instructions are required.

In the region-avoidance task, BehAV shows higher trajectory similarity in successful cases due to its cost-map design for non-traversable regions. InstructNav produces unstable paths because of its reliance on landmarks. NORM-Nav completes avoidance tasks consistently and achieves the best performance across all metrics.

In the traversable-obstacle task, BehAV and InstructNav fail to traverse regions misclassified as obstacles. INF successfully handles traversable obstacles but does not account for behavioral preferences, leading to degraded performance in combined scenarios. NORM-Nav overcomes these limitations by traversing misclassified obstacles while simultaneously following user-specified constraints, giving it a clear advantage.

In the combined task, users provide instructions that integrate all subtasks. NORM-Nav successfully executes these instructions and achieves superior results in all metrics. The generated trajectories are more efficient, and closer to human-operated trajectories compared with the baselines.

\subsection{Evaluation of Cost Distribution Shaping}
\begin{figure}[!t]
    \centering
    \includegraphics[width=0.48\textwidth]{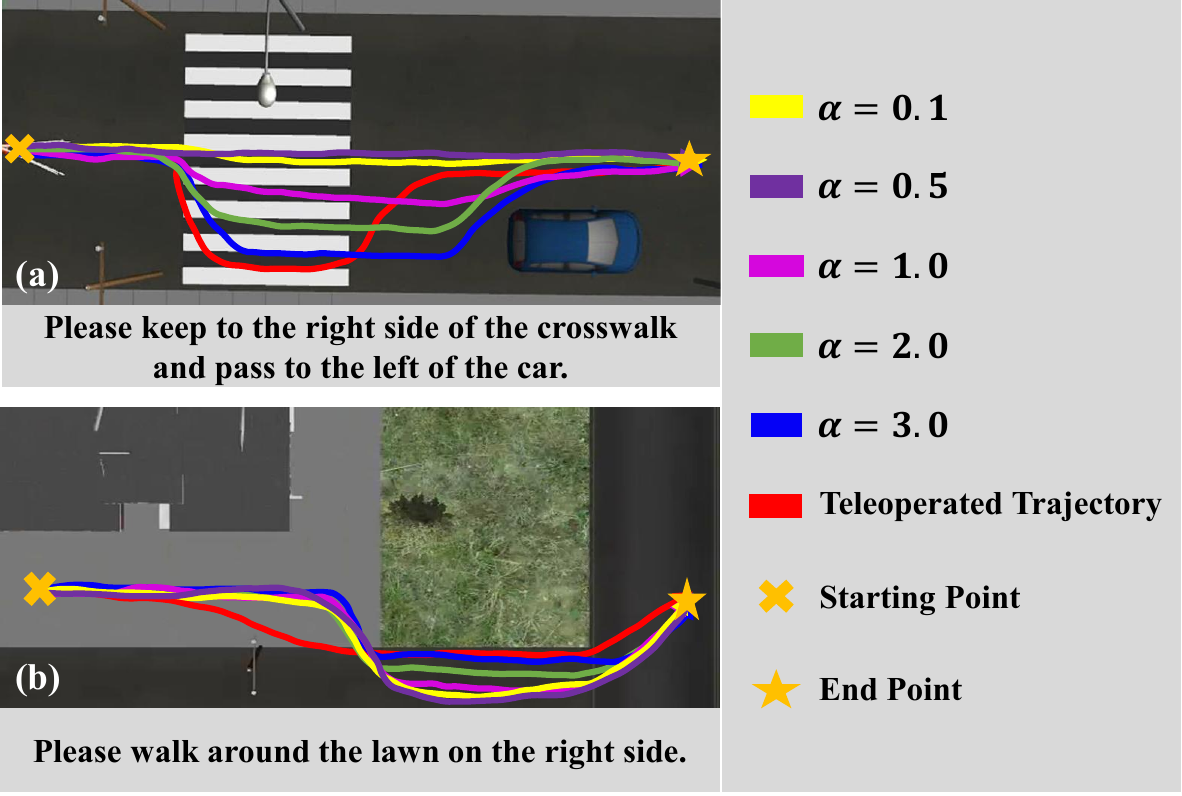}
    \caption{Effect of interpolation parameter $\alpha$ on trajectory generation. Larger $\alpha$ values lead to stronger compliance with directional preferences, while smaller values produce more conservative paths.}
    \label{gradient_exp}
    \vspace{-4mm}
\end{figure}

To evaluate the influence of the interpolation parameter $\alpha$ on navigation performance, we configured different $\alpha$ values in NORM-Nav. As defined in Eq.~\ref{eq:interpolation}, $\alpha$ determines the non-linear interpolation of cost values, thereby shaping the distribution of preference enforcement around objects. The experimental results are illustrated in Fig.~\ref{gradient_exp}.

When $\alpha < 1.0$, the robot trajectories exhibit weak adherence to the directional preferences. When $\alpha = 1.0$, the generated paths moderately reflect the intended constraints. When $\alpha > 1.0$, the robot follows the instructions more strictly. In summary, Fig.~\ref{gradient_exp} reveals that smaller $\alpha$ values favor conservative trajectories with greater obstacle clearance, while larger values enhance behavioral fidelity. Empirical observations suggest that settings with $\alpha > 1.0$ provide the most appropriate balance between safety and compliance with user-specified constraints.

\subsection{Evaluation of Speed Constraints}
The ability of the proposed method to enforce speed-related behavioral constraints is evaluated through instructions specifying three velocity modes: quickly, slowly, and normal (default), as illustrated in Fig.~\ref{teaser}. The results show that the robot adapts its translational velocity in real time according to the given commands: traversal speed increases in regions, decreases when instructed to move slowly near objects such as vehicles, and remains at a default pace when no explicit constraints are imposed. These observations indicate that natural language instructions are reliably translated into consistent low-level velocity control within the proposed method.

\subsection{Real-World Task Demonstrations}
We conduct representative real-world tests covering region following, region avoidance, traversable obstacle handling, and combined behaviors. Goal points are manually set, and behavioral constraints are given in natural language.

As shown in Fig.~\ref{real_exp}, the robot follows side-specific rules, such as keeping to the right of a road or bypassing pedestrians from the instructed side, and it reaches the goals without collisions. It also passes through a curtain that LiDAR misclassifies as non-traversable and avoids a manhole cover from the left, even though the cover is not detected as an obstacle.

These demonstrations confirm that the method reliably executes natural language instructions and generates safe, efficient, and human-like trajectories. In particular, the method correctly handles obstacles with ambiguous traversability semantics, such as curtains or manhole covers, which are prone to misclassification in conventional LiDAR-based navigation.

\section{Conclusion}

In this work, we present NORM-Nav, a zero-shot navigation framework that integrates natural language behavioral constraints into costmap-based planning. The framework parses free-form instructions with a language model, grounds them through vision–LiDAR perception, and encodes them into multi-layer costmaps so that planned trajectories remain both geometrically feasible and socially compliant. Experiments in simulation and real-world environments show that NORM-Nav achieves higher success rates, greater efficiency, closer adherence to behavioral rules, and stronger alignment with human-preferred paths than representative baselines. The results indicate that NORM-Nav provides a practical and generalizable solution for behavior-aware navigation.

\bibliographystyle{IEEEtran}  
\bibliography{ref}

\end{document}